\documentclass[conference]{IEEEtran}
\ifCLASSOPTIONcompsoc
  \usepackage[nocompress]{cite}
\else
  \usepackage{cite}
\fi
\usepackage{balance}

\ifCLASSINFOpdf
   \usepackage[pdftex]{graphicx}
   \graphicspath{{Figs/}}
  \usepackage{color, colortbl}
  \definecolor{Gray}{gray}{0.9}
\else
\fi
\begin{document}
\title{Assessing the Utility of Weather Data for Photovoltaic Power Prediction}



% compsoc conferences compared to traditional conferences), use this
\author{\IEEEauthorblockN{Reza Zafarani, Sara Eftekharnejad, and Urvi Patel}
\IEEEauthorblockA{Department of Electrical Engineering and Computer Science\\
Syracuse University\\
{\small \texttt{\{rzafaran, seftekha, uspatel\}@syr.edu}}}
}


\maketitle

\begin{abstract}
Photovoltaic systems have been widely deployed in recent times to meet the increased electricity demand as an environmental-friendly energy source. The major challenge for integrating photovoltaic systems in power systems is the unpredictability of the solar power generated. In this paper, we analyze the impact of having access to weather information for solar power generation prediction and find weather information that can help best predict photovoltaic power.  

\end{abstract}

% \fi
\IEEEpeerreviewmaketitle

\section{Introduction}
Renewable resources are now being deployed more than ever. In United States alone, 12.87\% of the total power generated in 2013 was from renewables. However, only 0.4\% of this quantity was from photovoltaic systems, mostly due to unpredictability in solar power. 

By harvesting Solar Power efficiently, the dependence on fossil fuels can be significantly reduced. To integrate renewables effectively into the power grid, a power generation estimation is necessary to manage grid load distribution. In photovoltaic power systems, the power generated is not easily predictable as it depends on various external parameters, especially weather conditions. 

Hence, it is of significant interesting to determine how accurately solar generation can be predicted by utilizing weather data. Consequently, it is important to determine the weather parameters that can help best predict the photovoltaic power generated. Numerous models have been proposed for solar power prediction with high accuracy rates. However, identifying weather factors which influence the solar power prediction most are less explored. In this paper, we investigate weather parameters that can help best predict solar power.

The rest of the paper is organized as follows: We first review models proposed to predict solar power generation in section 2. Then, in Section 3, we briefly review the dataset used in this study and proceed to identify weather factors affecting solar power generation. We conclude this paper in Section 4.

\section{Related Work}
Extensive literature exist on predicting photovoltaic power generation. The majority of the studies have employed data mining and machine learning algorithms such as Classification and Regression techniques. Examples include Support Vector Machines, Artificial Neural Networks, fuzzy networks, among other algorithms to forecast solar power generation. More commonly, one of two paths are taken by researchers. Either solar irradiance is predicted first and then based on the predicted value, power generation is calculated, or, the power generation is directly predicted.

\subsection{Predicting Solar Irradiance}

Most models proposed to predict Solar Irradiance are based on weather parameter values such as temperature, pressure, humidity, precipitation, wind direction and wind speed as they directly influence Solar Irradiance. Mellit and Pavan \cite{mellit201024} used mean solar irradiance and daily mean temperature values as parameters to predict 24 hour-ahead solar irradiance using Multilayer Perceptron (MLP) model with Back Propagation algorithm. In another model developed based on neural networks, Abadi et al.\cite{abadi2014extreme} estimate hourly solar radiation on horizontal surface. They used Extreme Learning Machine (ELM), with meteorological data such as temperature, humidity, wind direction and wind speed to train the model. 

Alanazi et al.\cite{alanazi2016long} proposed a different approach using Neural networks toolbox and Global Horizontal Irradiance (GHI) values as main factor for long term solar forecasting. The method involves a set of pre- and post-processes to be carried out before and after the forecast is obtained. Nomiyama et al.\cite{nomiyama2011study} proposed three methods for forecasting global solar radiation (GSR): Descriptive statistics, Factor analyses and Binary trees. They used weather forecast data along with clearness index which is calculated as a ratio of GSR and extra-terrestrial solar radiation for forecasting. Based on weather information, these methods forecast two-day-ahead, one-day-ahead and three-hour-ahead GSR, respectively. 

Mori and Takahashi \cite{mori2012data} proposed a data mining method based on CART (Classification and Regression Trees) for selecting explanatory variables for global solar radiation forecasting. The Variable importance of CART is used as an index to prioritize variable. Among all the variables they considered, their study found Solar Radiation, duration of sunlight and altitude of sun to be more important in both Winter and Summer. Humidity and Temperature also had high importance in Summer. 

\subsection{Predicting Solar Power}

There are various models proposed that directly predict the solar power generated. Most of these models also utilize weather information or solar irradiance values. Ding et al. \cite{ding2011ann} proposed a model that follows this approach. They used a Feed-forward neural network (FNN) with an improved back-propagation algorithm along with a similar day selection algorithm based on forecast day information. The similar day selection algorithm selected a day from historic data that has the most similar weather conditions as the forecast day. The model also considered high, low and average temperatures on the selected day as well as predicted values on forecast day. Their study shows that forecasts under different weather types can be carried out using only historical power data and weather data. 

Zhang et al.\cite{zhang2015day} also presented a similar day-based forecasting tool for 24 hour-ahead forecasting for a small scale solar power output. Their proposed forecasting method had two components: Similar Day Detection (SDD) engine and forecasting engine. Their study found the most effective external variable as the irradiance forecasts for all geographical locations that was studied. As weather forecasts are used as model inputs, inaccuracy in those forecasts lead into inevitable solar power forecast errors.

Sharma et al.\cite{sharma2011predicting} proposed a model for automatically creating site-specific prediction models for solar power generation using weather forecast data as input to the model. They used observational solar intensity data from an extended weather station rooftop deployment to build the model. Their study compared the performance of Linear least squares regression and support vector machines using linear kernel, polynomial kernel, and RBF kernel and found the SVM with RBM kernel to be the most accurate. Kang et al.\cite{kang2011development} attempted to forecast PV power by developing an algorithm based on rainfall patterns in the past 5 years' data. They clustered this dataset and ultimately matched the same date over 5 years to find a pattern. 

Shi et al.\cite{shi2012forecasting} proposed algorithms to forecast power output of photovoltaic systems based on weather classification and Support Vector Machines (SVM). They classified weather data based on four weather conditions: clear sky, cloudy day, foggy day, rainy day. They created a model for one-day-ahead PV power output forecasting for a single station based on the weather forecasting data, actual historical power output data, and the SVMs. 

In a hybrid approach proposed by Mandal et al.\cite{mandal2012forecasting}, the authors forecast one-hour-ahead power output of a PV system using a combination of wavelet transform (WT) and RBFNN, along with solar radiation and temperature data. 
Xu et al.\cite{xu2012short} adopted a weighted Support Vector Machine (WSVM) to forecast the short-term PV power. The proposed model selected 5 days with the most similarity to the day to be forecasted as the training samples.

\section{Photovoltaic Power Forecasting\\ with Weather Data}

We perform regression on a real-world PV data to determine how accurately PV power can be estimated by exploiting weather information. We find weather factors that can best predict the Photovoltaic Power. We first describe the data used for PV forecasting. Next, we discuss experiments conducted using this datasets. The experiments employ two regression techniques: (1) Least Absolute Shrinkage and Selection Operator (LASSO)\cite{tibshirani1996regression} and (2) Linear Regression.

\subsection{Data}

We analyze a real-world dataset provided by Greenview Energy Management Systems. The data consists of more than 100 parameters collected during 29 June, 2016 to 25 February, 2017 from the solar panels planted in Syracuse, NY. 

The dataset consists of three different types of data: 
\begin{enumerate}
\item \textbf{Weather Data.} This part of our data consists of different weather parameters measured such as temperature, cloud cover, humidity, precipitation, pressure, visibility, wind direction and wind speed. It  consists of predicted values of these measures for five following days.
\item\textbf{Meter Data.} The second part is meter data, which consists of parameters of the PV system including PV power which we are trying to forecast.
\item\textbf{Solar Radiation Data.} Final portion is Solar Radiation data, which consists of the Minimum, Maximum, Average and Instantaneous values of Solar Irradiance.   

\end{enumerate} 

\subsection{Predicting Solar Power}

To predict solar power, we utilize two regression techniques: LASSO and Linear Regression. We predict PV power using weather data by performing  LASSO and Linear Regression on the data. LASSO performs both variable selection and regularization. We evaluate the results using Mean Squared Error (MSE), which is a measure of the difference between actual values and predicted values. The lower the MSE, the more accurate the algorithm. 

\subsubsection{Prediction}
We first perform LASSO to investigate the feasibility of predicting PV power using weather data. Performing LASSO on the dataset with default parameters, we obtain an MSE value of 5.5436. Constant $\alpha$ in LASSO's mathematical formulation affects the accuracy of the model. To find the optimal value of $\alpha$, we perform LASSO with Cross Validation with 10 folds and the value of $\alpha$ obtained was 0.1722. Training the model using this $\alpha$ value and 60\% sampled data, the MSE value obtained was 5.5045. We tested different values of $\alpha$ and obtained the corresponding MSE values as shown in the Table \ref{table_lasso}.

\begin{table}[h]
\renewcommand{\arraystretch}{1.5}
 %if using array.sty, it might be a good idea to tweak the value of
 %\extrarowheight as needed to properly center the text within the cells
\caption{MSE values for different $\alpha$}
\label{table_lasso}
\centering
% Some packages, such as MDW tools, offer better commands for making tables
% than the plain LaTeX2e tabular which is used here.
\begin{tabular}{|c|c|}
\hline
$\alpha$ & MSE\\
\hline
0.1722 & 5.5045\\

0.15 & 5.4959\\

0.10 & 5.4739\\

0.05 & 5.4555\\

0.001 & 5.4459\\

\hline

\end{tabular}
\end{table}

LASSO allows us to obtain most important weather parameters, i.e., perform \textit{feature selection}. Features are the parameters which are used to predict a variable. In our experiments, we used all the weather parameters as features. we selected features with non-zero coefficient. We perform LASSO again on the selected features using $\alpha$ as 0.1722, for which we got the MSE values of 5.5045. This value is the same as the MSE value obtained using all the features. Thus we can effectively use LASSO for feature selection without affecting accuracy. 

\begin{figure}[h]
\centering
\includegraphics[width=0.5\textwidth]{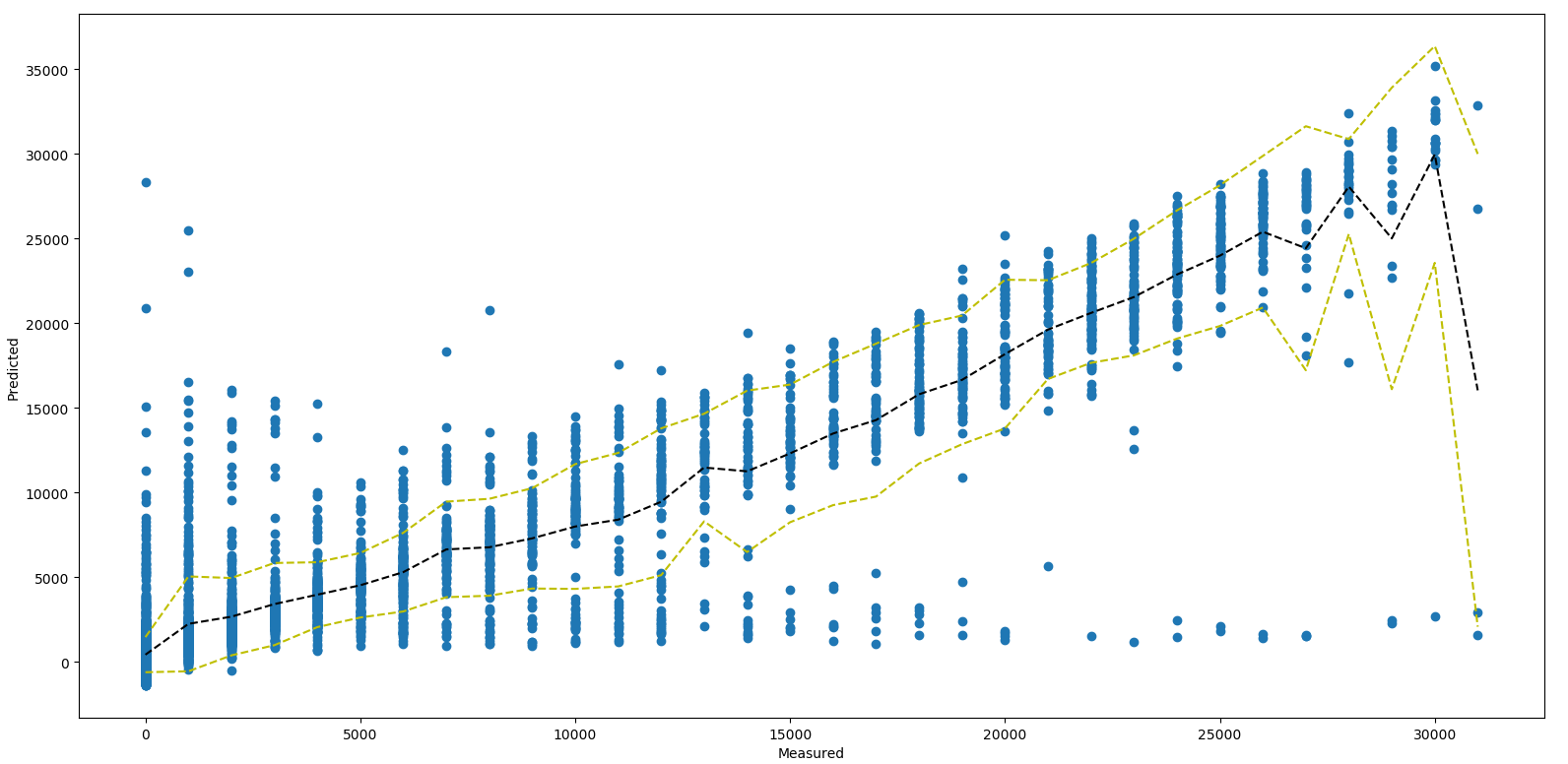}
 %where an .eps filename suffix will be assumed under latex, 
 %and a .pdf suffix will be assumed for pdflatex; or what has been declared
 %via \DeclareGraphicsExtensions.
\caption{Predicted vs Measured value of PV power}
\label{fig_pred_meas}
\end{figure}

Next, we perform Linear Regression on the whole dataset and obtain the MSE value of 5.4248. We perform Linear Regression on 60\% of data and tested the model on other 40\% of data. The MSE value obtained in this case was 5.4147. Since Linear Regression performed better than LASSO, we selected this regression method for the following experiments. To visualize the performance of Linear Regression, we plotted the predicted PV power values versus measured values as shown in Figure \ref{fig_pred_meas}. Here, we scaled the values from kW to W and then grouped the measured values so that we can better visualize the variation in predicted values. 

\subsubsection{Feature Importance Analysis}   
To find the most important features, we use the absolute coefficient value of each feature obtained through Linear Regression and selected the top 25 features with highest coefficient values. Performing Linear Regression on these selected features, we obtained the MSE value of 5.4968. We conductance importance analysis for top 25 features. We perform Linear Regression on 25 features repeatedly leaving one feature at a time. The Table \ref{table_lr1} shows the MSE value obtained when the feature was opted out. 

\begin{table}[h]
\renewcommand{\arraystretch}{1.5}
 %if using array.sty, it might be a good idea to tweak the value of
 %\extrarowheight as needed to properly center the text within the cells
\caption{Importance Analysis of Weather Features}
\label{table_lr1}
\centering
% Some packages, such as MDW tools, offer better commands for making tables
% than the plain LaTeX2e tabular which is used here.

\begin{tabular}{|c|c|c|}
\hline
k & Feature & MSE\\
\hline
1 & $day4precipMM$ & 5.5108\\

2 & $day2precipMM$ & 5.5007\\

3 & $day1precipMM$ & 5.5057\\

4 & $day5precipMM$ & 5.4989\\

5 & $day3precipMM$ & 5.4970\\

\rowcolor{Gray}
6 & $input1A$ & 8.8404\\

7 & $precipMM$ & 5.4969\\

8 & $visibility$ & 5.5036\\

9 & $windspeedMiles$ & 5.4972\\

10 & $day1windspeedMiles$ & 5.5036\\

11 & $input1B$ & 5.5163\\
\rowcolor{Gray}
12 & $day1tempMaxF$ & 5.5667\\

13 & $day4tempMaxF$ & 5.5127\\

14 & $input1D$ & 5.5130\\

15 & $windspeedKmph$ & 5.4971\\

16 & $day5windspeedMiles$ & 5.4980\\

17 & $day4tempMinF$ & 5.5072\\

18 & $day4windspeedMiles$ & 5.4966\\

19 & $day2windspeedMiles$ & 5.4963\\

20 & $day3windspeedMiles$ & 5.4974\\

21 & $humidity$ & 5.4992\\

22 & $day2tempMinF$ & 5.5140\\

23 & $day5tempMaxF$ & 5.4992\\

24 & $day3tempMaxF$ & 5.4998\\

25 & $day5tempMinF$ & 5.4969\\

\hline

\end{tabular}
\end{table}
  
\begin{figure}[h]
\centering
\includegraphics[width=0.5\textwidth]{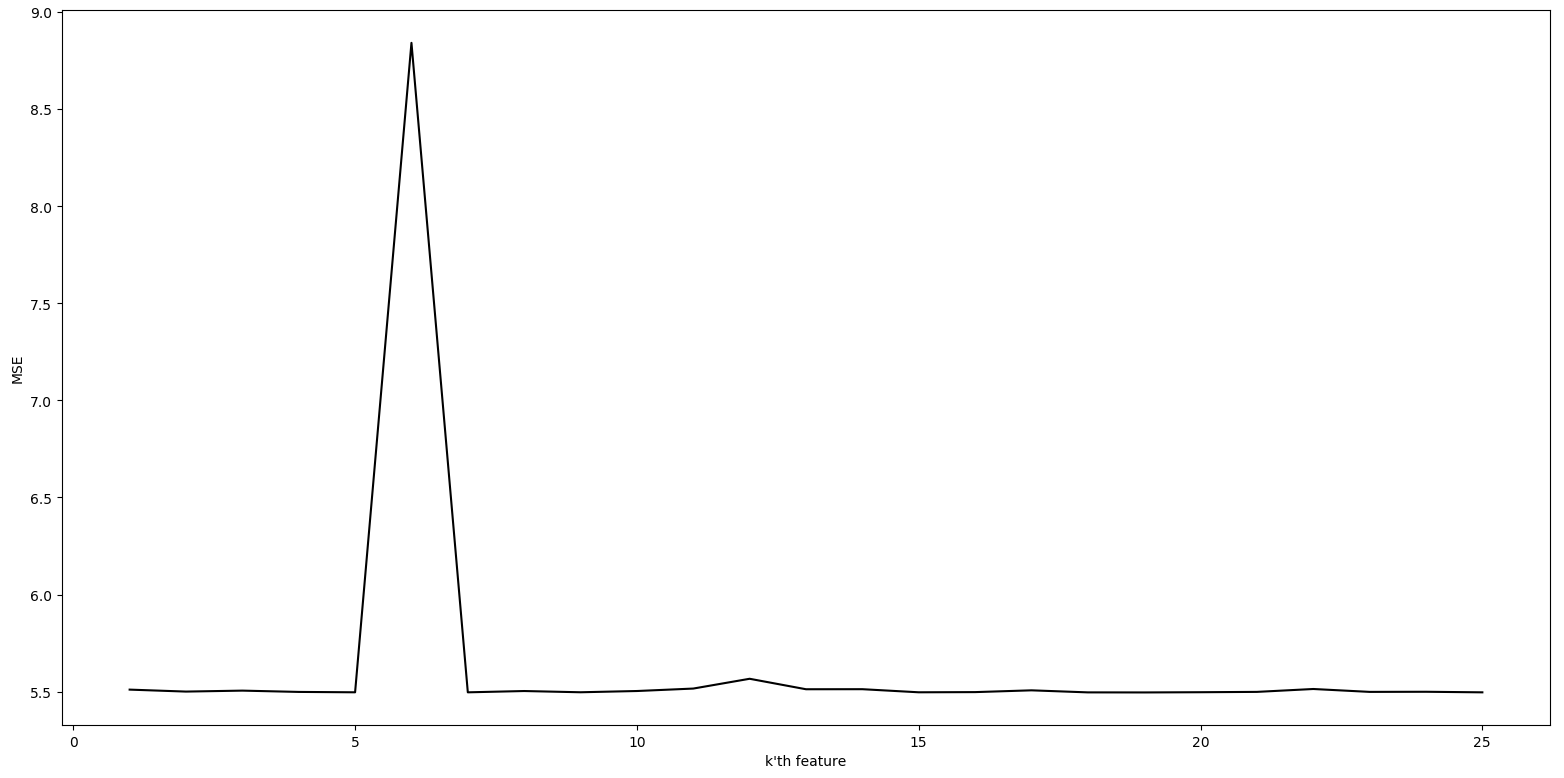}
\caption{Importance Analysis: MSE vs $k$}
\label{fig_knockouts}
\end{figure}

The highlighted rows in the table show the features that when opted out, the MSE value increases. Consequently we can say that these are more important features. The feature 'input1A' is the instantaneous solar irradiance value and 'day1tempMaxF' is the predicted maximum temperature in Fahrenheit one day following the forecast day. Figure \ref{fig_knockouts} plots these MSE values for each $k$.

\subsubsection{Limited Features Analysis}

\begin{figure}[h]
\centering
\includegraphics[width=0.5\textwidth]{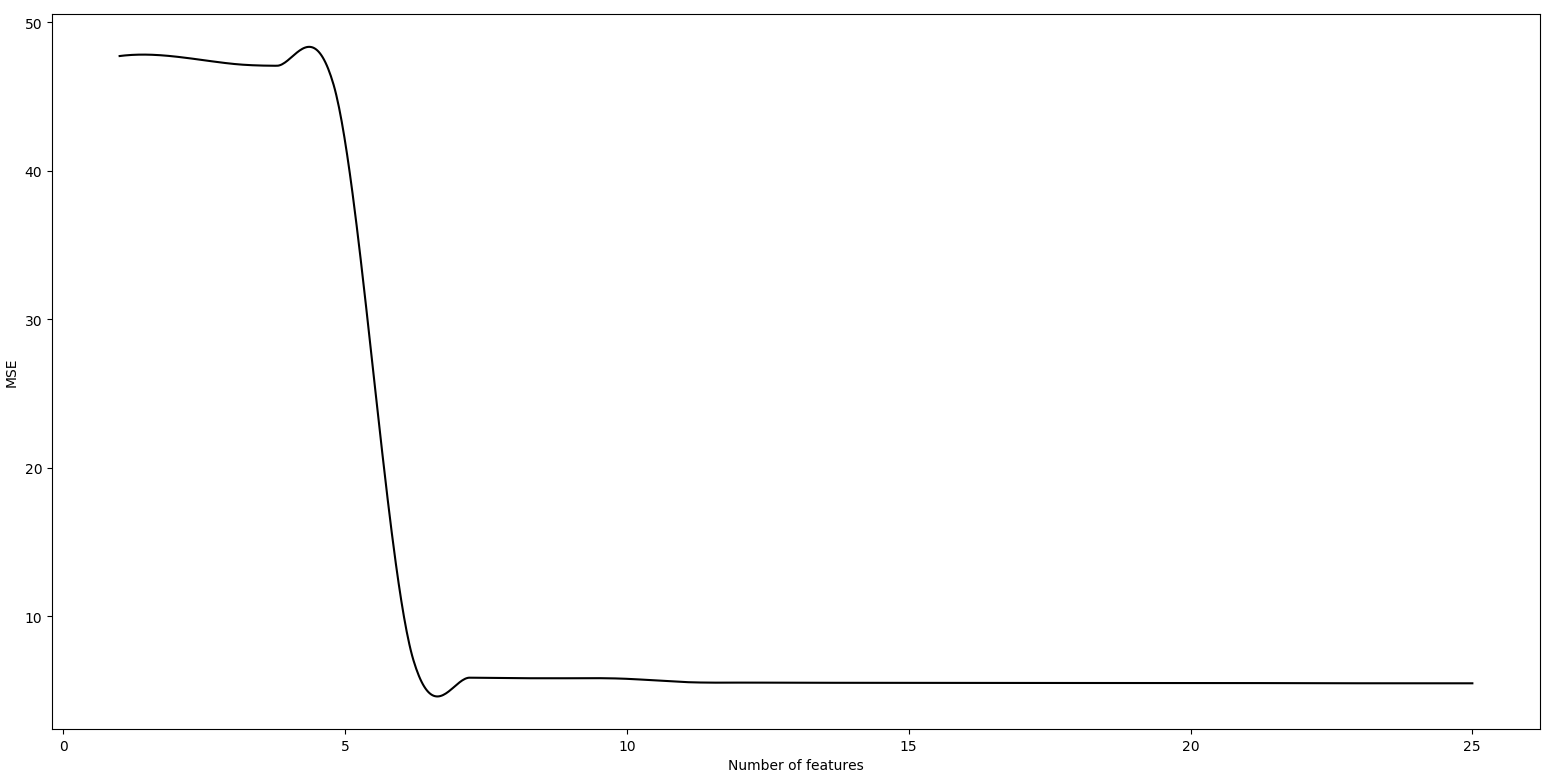}
 %where an .eps filename suffix will be assumed under latex, 
 %and a .pdf suffix will be assumed for pdflatex; or what has been declared
 %via \DeclareGraphicsExtensions.
\caption{MSE versus Number of Weather Features ($k$)}
\label{fig_mse_k}
\end{figure}

In practical scenarios, many of the weather features might not be available or performing the prediction using all the available features might be computationally prohibitive. In such cases, to find the optimal number of weather features to be selected, we performed Linear Regression repeatedly on the data using the top $k$ = 1, 2, 3...,30 features.   Figure \ref{fig_mse_k} plots the MSE values versus the number of features selected ($k$). As can be seen, after $k=6$, the MSE saturates. This observation indicates that we can forecast PV power accurately using only top 6 features.

\subsubsection{Limited Data Analysis}   

\begin{figure}[h]
\centering
\includegraphics[width=0.5\textwidth]{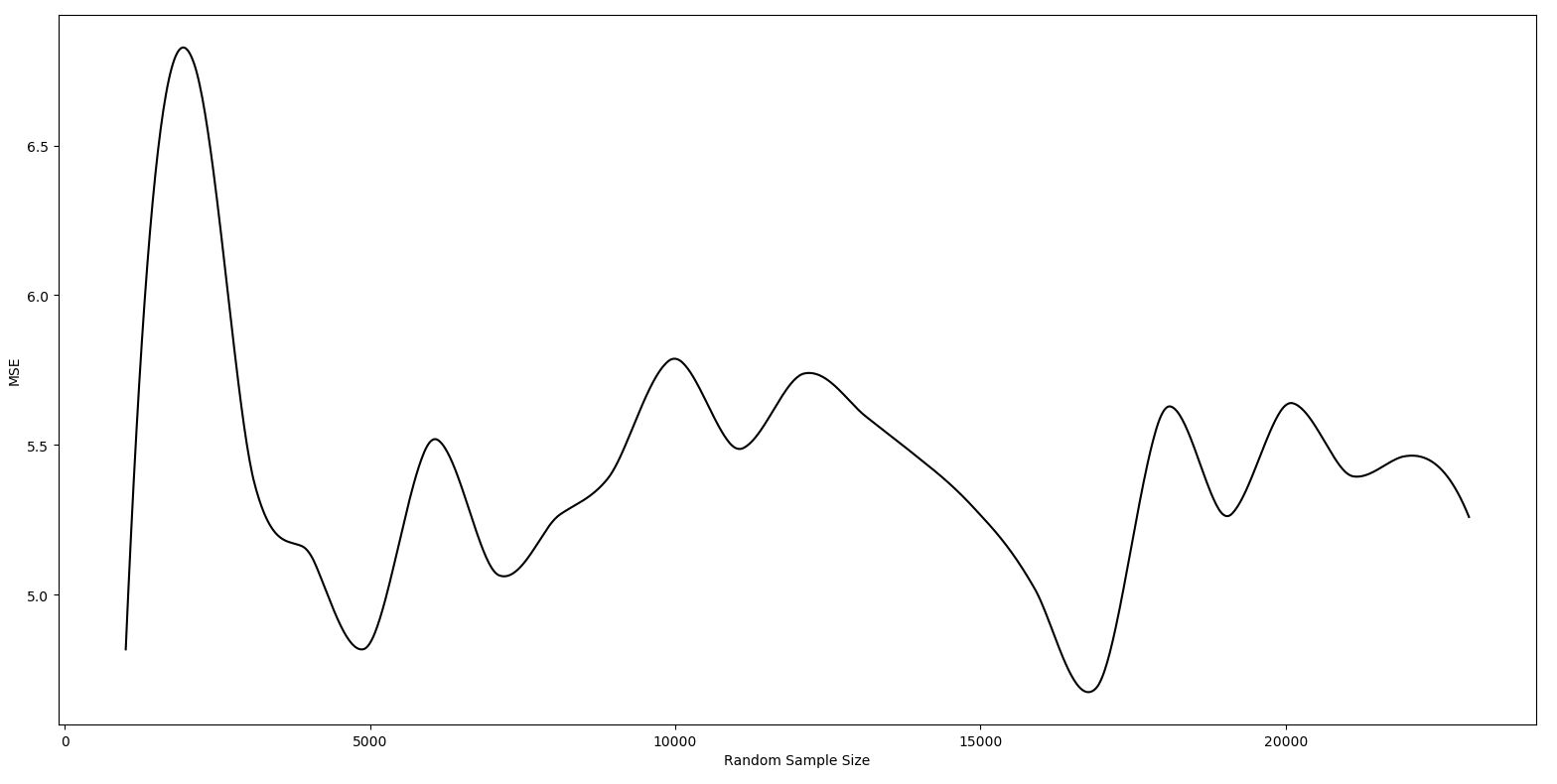}
\caption{MSE versus Random Sample Size}
\label{fig_partialdata}
\end{figure}

Next we try to find out how much weather data is required. We measure how weather data size influences the MSE value. In particular, we aim to determine with how much weather data can one provide reasonably accurate predictions on solar power generation. To conduct this experiment, we took several random samples from the data and measure the MSE value. Figure \ref{fig_partialdata} plots the sample size versus the MSE value. As we can see from the figure, the fluctuations in MSE values are considerably less after sample size of about 5000, which is about 2 months of data.

\section{Conclusion}
To find the features that can best predict photovoltaic power, we found that Linear Regression performs better than LASSO. Using Linear Regression, we identified top 25 important features and found Instantaneous Solar Irradiance to be the most important. We found that only top 6 features are sufficient to predict solar power with reasonable accuracy using limited features analysis. Next we found that only about 2 months data is needed to train the model without affecting accuracy.





\balance
\bibliographystyle{IEEEtran}
\bibliography{IEEEabrv,bibfile}

\begin{thebibliography}{10}
\providecommand{\url}[1]{#1}
\csname url@samestyle\endcsname
\providecommand{\newblock}{\relax}
\providecommand{\bibinfo}[2]{#2}
\providecommand{\BIBentrySTDinterwordspacing}{\spaceskip=0pt\relax}
\providecommand{\BIBentryALTinterwordstretchfactor}{4}
\providecommand{\BIBentryALTinterwordspacing}{\spaceskip=\fontdimen2\font plus
\BIBentryALTinterwordstretchfactor\fontdimen3\font minus
  \fontdimen4\font\relax}
\providecommand{\BIBforeignlanguage}[2]{{%
\expandafter\ifx\csname l@#1\endcsname\relax
\typeout{** WARNING: IEEEtran.bst: No hyphenation pattern has been}%
\typeout{** loaded for the language `#1'. Using the pattern for}%
\typeout{** the default language instead.}%
\else
\language=\csname l@#1\endcsname
\fi
#2}}
\providecommand{\BIBdecl}{\relax}
\BIBdecl

\bibitem{mellit201024}
A.~Mellit and A.~M. Pavan, ``A 24-h forecast of solar irradiance using
  artificial neural network: Application for performance prediction of a
  grid-connected pv plant at trieste, italy,'' \emph{Solar Energy}, vol.~84,
  no.~5, pp. 807--821, 2010.

\bibitem{abadi2014extreme}
I.~Abadi, A.~Soeprijanto \emph{et~al.}, ``Extreme learning machine approach to
  estimate hourly solar radiation on horizontal surface (pv) in surabaya-east
  java,'' in \emph{Information Technology, Computer and Electrical Engineering
  (ICITACEE), 2014 1st International Conference on}.\hskip 1em plus 0.5em minus
  0.4em\relax IEEE, 2014, pp. 372--376.

\bibitem{alanazi2016long}
M.~Alanazi, A.~Alanazi, and A.~Khodaei, ``Long-term solar generation
  forecasting,'' in \emph{Transmission and Distribution Conference and
  Exposition (T\&D), 2016 IEEE/PES}.\hskip 1em plus 0.5em minus 0.4em\relax
  IEEE, 2016, pp. 1--5.

\bibitem{nomiyama2011study}
F.~Nomiyama, J.~Asai, T.~Murakami, and J.~Murata, ``A study on global solar
  radiation forecasting using weather forecast data,'' in \emph{Circuits and
  Systems (MWSCAS), 2011 IEEE 54th International Midwest Symposium on}.\hskip
  1em plus 0.5em minus 0.4em\relax IEEE, 2011, pp. 1--4.

\bibitem{mori2012data}
H.~Mori and A.~Takahashi, ``A data mining method for selecting input variables
  for forecasting model of global solar radiation,'' in \emph{Transmission and
  Distribution Conference and Exposition (T\&D), 2012 IEEE PES}.\hskip 1em plus
  0.5em minus 0.4em\relax IEEE, 2012, pp. 1--6.

\bibitem{ding2011ann}
M.~Ding, L.~Wang, and R.~Bi, ``An ann-based approach for forecasting the power
  output of photovoltaic system,'' \emph{Procedia Environmental Sciences},
  vol.~11, pp. 1308--1315, 2011.

\bibitem{zhang2015day}
Y.~Zhang, M.~Beaudin, R.~Taheri, H.~Zareipour, and D.~Wood, ``Day-ahead power
  output forecasting for small-scale solar photovoltaic electricity
  generators,'' \emph{IEEE Transactions on Smart Grid}, vol.~6, no.~5, pp.
  2253--2262, 2015.

\bibitem{sharma2011predicting}
N.~Sharma, P.~Sharma, D.~Irwin, and P.~Shenoy, ``Predicting solar generation
  from weather forecasts using machine learning,'' in \emph{Smart Grid
  Communications (SmartGridComm), 2011 IEEE International Conference on}.\hskip
  1em plus 0.5em minus 0.4em\relax IEEE, 2011, pp. 528--533.

\bibitem{kang2011development}
M.-C. Kang, J.-M. Sohn, J.-y. Park, S.-K. Lee, and Y.-T. Yoon, ``Development of
  algorithm for day ahead pv generation forecasting using data mining method,''
  in \emph{Circuits and Systems (MWSCAS), 2011 IEEE 54th International Midwest
  Symposium on}.\hskip 1em plus 0.5em minus 0.4em\relax IEEE, 2011, pp. 1--4.

\bibitem{shi2012forecasting}
J.~Shi, W.-J. Lee, Y.~Liu, Y.~Yang, and P.~Wang, ``Forecasting power output of
  photovoltaic systems based on weather classification and support vector
  machines,'' \emph{IEEE Transactions on Industry Applications}, vol.~48,
  no.~3, pp. 1064--1069, 2012.

\bibitem{mandal2012forecasting}
P.~Mandal, S.~T.~S. Madhira, J.~Meng, R.~L. Pineda \emph{et~al.}, ``Forecasting
  power output of solar photovoltaic system using wavelet transform and
  artificial intelligence techniques,'' \emph{Procedia Computer Science},
  vol.~12, pp. 332--337, 2012.

\bibitem{xu2012short}
R.~Xu, H.~Chen, and X.~Sun, ``Short-term photovoltaic power forecasting with
  weighted support vector machine,'' in \emph{Automation and Logistics (ICAL),
  2012 IEEE International Conference on}.\hskip 1em plus 0.5em minus
  0.4em\relax IEEE, 2012, pp. 248--253.

\bibitem{tibshirani1996regression}
R.~Tibshirani, ``Regression shrinkage and selection via the lasso,''
  \emph{Journal of the Royal Statistical Society. Series B (Methodological)},
  pp. 267--288, 1996.

\end{thebibliography}
%\bibliography{bibfile}

%\begin{thebibliography}{1}

%H.~Kopka and P.~W. Daly, \emph{A Guide to \LaTeX}, 3rd~ed.\hskip 1em %plus

%\end{thebibliography}

\end{document}